\documentclass[letterpaper, 10 pt, conference]{ieeeconf}  

\IEEEoverridecommandlockouts                              

\usepackage{graphics} 
\usepackage{epsfig} 
\usepackage{mathptmx} 
\usepackage{times} 
\usepackage{amsmath} 
\usepackage{amssymb}  
\usepackage{mathtools}
\usepackage{algorithm}
\usepackage[noend]{algpseudocode}
\usepackage{threeparttable, type1cm}
\usepackage{graphicx, textcomp, xcolor}

\title{\LARGE \bf
Multi-Robot Patrol Algorithm with Distributed Coordination and Consciousness of the Base Station's Situation Awareness
}

\author{Kazuho Kobayashi$^{1}$, Seiya Ueno$^{2}$, and Takehiro Higuchi$^{2}$
\thanks{$^{1}$Kazuho Kobayashi is with Department of Mechanical Engineering, Materials Science, and Ocean Engineering, Graduate School of Engineering Science, Yokohama National University, Yokohama-city, Kanagawa, Japan
        {\tt\small kobayashi-kazuho-dj@ynu.jp \textit{or} kazuho.kobayashi.ynu@gmail.com}}%
\thanks{$^{2}$Seiya Ueno and Takehiro Higuchi with Faculty of Environment and Information Sciences, Yokohama National University
        {\tt\small \{ueno-seiya-wk, higuchi\}@ynu.ac.jp}}%
\thanks{\textbf{This work has been submitted to the IEEE for possible publication. Copyright may be transferred without notice, after which this version may no longer be accessible.}}
}

\begin{document}

\maketitle
\thispagestyle{empty}
\pagestyle{empty}

\begin{abstract}
  Multi-robot patrolling represents a potential application for robotic systems to survey wide areas efficiently. However, such systems have seen few examples of real-world applications due to their lack of human predictability. This paper proposes the Local Reactive (LR) algorithm for multi-robot patrolling, aiming to satisfy two primary needs: (i) efficient patrolling and (ii)providing humans with enhanced situation awareness to improve system predictability. Each robot selects its patrol target from the local areas around its current location based on two requirements: (i)patrolling locations with greater need and (ii)reporting its achievements to the base station. The algorithm is distributed and coordinates the robots without centralized control, enabling them to share their patrol achievements and the degree of need to report to the base station. The proposed algorithm outperformed existing algorithms in both patrolling effectiveness and the base station’s situation awareness.
\end{abstract}

\section{INTRODUCTION}
Multi-robot patrol missions involve multiple mobile unmanned systems surveying an assigned area to detect undesirable situations. Potential applications may include patrolling disaster zones, large recreational sites, or border regions.

Despite the potential, there are few examples of real-world applications of such systems, often attributed to their lack of human predictability~\cite{Schranz2020}. Since the robots work remotely and their interactions with other robots are invisible, the system’s output is often unpredictable, resulting in concern about robot malfunctions and harmful behaviors~\cite{Carrillo-Zapata2020, Gans2021}.

A possible approach to mitigate this issue involves frequently reporting mission progress and system status to the base station, thereby enhancing situation awareness (SA)~\cite{Endsley1988}. SA may improve system predictability by enabling the base station to grasp the mission progress, detect system malfunctions, and intervene to support robots’ decision-making.

Various approaches to multi-robot patrolling have been explored in the literature. Banfi et al. proposed the Situation-Aware Patroling, which considers both area coverage and the latency of information delivery~\cite{Banfi2015}. Similarly, Scherer and Rinner also developed an algorithm that computes routes for both patrolling and information delivery~\cite{Scherer2020}. However, such algorithms depend on central, preliminary computations for patrolling. Distributed and reactive algorithms~\cite{Huang2019} are desirable due to their robustness, scalability, flexibility in mission configuration, and ease of deployment. Within this category, Machado et al. introduced the Conscientious Reactive (CR) algorithm~\cite{Machado2003}. More recent developments include algorithms that facilitate coordination among robots~\cite{Yan2016, Farinelli2017}. 

This study builds upon these efforts, developing a patrol algorithm conscious of the base station’s situation awareness. In the proposed Local Reactive (LR) algorithm, each robot selects its patrol target from the local area around itself based on two needs: patrolling and reporting to the base station. The following sections describe the problem definition (Section~\ref{Prob}), proposed algorithm (Section~\ref{Algo}), simulation results(Section~\ref{Sim}), and conclusion (Section~\ref{Conclude}).

\section{PROBLEM SETTINGS}\label{Prob}
The research assumes patrol for large open areas by unmanned aerial / surface / underwater vehicles (UAV / USV / UUV).

\subsection{Mission}
The scope of this research focuses on a reactive approach. Here, \textit{reactive} means that each robot, independent of others and the base station, determines its next patrol target upon completing its current target, without engaging in optimization calculations or preliminary planning for the global optima. This study defines the mission area as a two-dimensional grid map $G \coloneqq \{g^k\ |\ k = 1,2,\ldots ,K\}$. $g^k$ denotes each grid, corresponding to each patrol location, while $\textit{\textbf{x}}^k$ represents the location of the grid center, and \textit{K} is the number of grids. 

The parameter: \textit{idleness} $i^k$, associated with $g^k$~\cite{Machado2003}, represents the elapsed time since a robot has last visited the grid. For instance, when a robot visited the grid $g^k$ at time $t^k$, and no other robot visited since, idleness at time $t$ is computed as $i^k(t) = t - t^k$. The study discretizes time as integer values equal to or greater than zero.

Reactive patrolling is regarded as iterations of behaviors whereby individual robots traverse from one grid to another. Each robot progresses toward the center of its patrol target grid, and the period during which the robot is located inside a grid is considered the time required to patrol it.

\subsection{Robots}
A group of robots $R \coloneqq \{r_n\ |\ n=1,2, \ldots ,N\}$, patrols the field, where \textit{N} represents the number of robots. Each robot is capable of: (i)sensing its accurate location, $\textit{\textbf{x}}_n$; (ii)identifying other robots and establishing connection within range $d_s$; and (iii)maintaining the connection once established, provided the distance between the robots does not exceed range $d_c\ (>d_s)$. Robot $r_1$ serves as the base station, is fixed at the map's origin: $\textit{\textbf{x}}_1=(0,\ 0)^T$, and does not participate in patrolling.

As for the individual robot, this research assumes a small robot capable of moving forward and turning, such as a quad-rotor drone maintaining a constant altitude or a differential-driven ground vehicle.

\subsection{Performance metric}
This study introduces two metrics: \textit{graph idleness} ($I_G$) and \textit{worst idleness} ($I_W$). The instantaneous graph idleness $I_G(t_c)$ at time $t_c$ is computed as the mean value of $i^k(t)$ across all grids from the mission's start ($t = t_0$) to $t=t_c$ using the following equation:
\begin{equation}
  I_G(t_c) = \frac{1}{(t_c - t_0 + 1)K}\sum_{t=t_0}^{t_c}\sum_{k=1}^{K}i^k(t) 
\end{equation}

\noindent Finally, $I_G$ can be defined as $I_G \coloneqq I_G(T)$, where \textit{T} denotes the time at the mission's end. Instantaneous worst idleness $I_W(t_c)$ represents the largest value of $i^k(t)$ in the interval $t_0 \leq t \leq t_c$ as expressed by:
\begin{equation}
  I_W(t_c) = \mathop{\mathrm{max}}_{t_0 \leq t\leq t_c, k \in K}i^k(t)
\end{equation}

\noindent $I_W$ is then defined as $I_W \coloneqq I_W(T)$. $I_G$ and $I_W$ signify the general patrol performance and patrol's comprehensiveness, respectively. 

This study also introduces metrics to quantify the base station's SA: mean SA delay ($D_{\mathrm{MSA}}$) and worst SA delay ($D_{\mathrm{WSA}}$). The definitions follow an existing study~\cite{Kobayashi2023a} and provide insight into the timeliness with which the base station acquires information of each grid.

\section{PROPOSED ALGORITHM}\label{Algo}
\subsection{Robotic assumption and its sharing}\label{bot-assume}
Since idleness is not measurable by external sensors, $r_n \in R$ maintains an \textit{assumed idleness} $H_n(t) \coloneqq \{h_n^k(t)=[i_n^k(t),\ t_n^k(t)]\ |\ \forall k \in K\}$, where $i_n^k(t)$ is the assumed value of $i^k(t)$ by $r_n$ at time \textit{t} and $t_n^k(t)$ is the refresh time of $i_n^k(t)$. $r_n$ increments its $i_n^k(t)$ as time elapses based on the idleness definition, assuming no other robots visit $g^k$. The $t_n^k(t)$ remains constant during this procedure since the growth of $i_n^k(t)$ is not confirmed by observation. When $r_n$ has just visited $g^k$ at $t$, it updates the assumption as: $i_n^k(t) \leftarrow 0$ and $t_n^k(t) \leftarrow t$. 

Additionally, $r_n$ updates $H_n(t)$ upon connecting with other robots, following a procedure identical to \cite{Kobayashi2023a}. In brief, $r_n$ exchanges all $h_n^k(t) \in H_n(t)$ with $\forall r_m \in \mathcal{A}_n(t)$ at every timestep, wherein $\mathcal{A}_n(t)$ represents a group of robots directly connected to $r_n$. When $t_n^k(t)$ is smaller (i.e., the information is older) than $t_m^k(t)$, $r_n$ replaces $h_n^k(t)$ with the value of $h_m^k(t)$. $h_n^k(t)$ reflects $r_n$'s mission achievements, and it propagates through all other robots, including base station, as shared knowledge.

\subsection{Target selection}\label{tar_select}
$r_n \in R$ determines its patrol target $g_n^{\tau}(t)$ utilizing $H_n(t)$ and a parameter: $\epsilon_n(t)$, which signifies the communication necessity with the base station as delineated in Algorithm~\ref{alg1}. In brief, $r_n$ updates $H_n(t)$ following the procedure from Section~\ref{bot-assume} (line 2). If $r_n$ is connected to the base station, it resets $\epsilon_n(t)$ as zero, and logs the time as $c^b_n(t)$ (lines 5-6). Given that patrol advances over time and the reporting need to the base station increases accordingly, $\epsilon_n(t)$ grows over time (line 6) while it is capped by $\epsilon_{\mathrm{max}}$. The function \textit{update\_epsilon} further updates $\epsilon_n(t)$ (line 8). 

\begin{algorithm}[tbp]
  \caption{Target selection by $r_n$ at time $t$}
  \label{alg1}
  \begin{algorithmic}[1]
  \Procedure{target selection}{$H_n(t-1), g_n^{\tau}(t-1), \epsilon_n(t-1)$}
    \State $H_n(t) \leftarrow$ update\_assumption($H_n(t-1)$)
    \State $g_n^{\tau}(t) \leftarrow g_n^{\tau}(t-1)$
    \If{connected to the base station}
        \State $\epsilon_n(t) \leftarrow 0$
        \State $c_n^b(t) \leftarrow t$
    \EndIf
    \State $\epsilon_n(t) \leftarrow \operatorname{min}\{\epsilon_{max},\ \epsilon(t-1) + 1\}$
    \State $\epsilon_n(t) \leftarrow \mathrm{update\_epsilon}(r_n, mode)$
    \If{$r_n$ has just visited $g_n^{\tau}(t)$}
      \State $h_n^{\tau}(t) \leftarrow [0, t]$
      \State $g_n^{\tau}(t) \leftarrow \underset{g^k \in G_n^{\delta}(t)} {\operatorname{argmax}}\ U_n^k(t)$
      \State $h_n^{\tau}(t) \leftarrow [0, t]$
    \EndIf
    \State \textbf{return} $H_n(t), g_n^{\tau}(t), \epsilon_n(t)$
  \EndProcedure
  \end{algorithmic}
\end{algorithm}

Upon visiting its current target, $r_n$ selects a new target $g_n^{\tau}(t)$ after updating the assumption about the current target as \textit{visited} (lines 9-11). $r_n$ considers a set of grids $G_n^{\delta}(t) \coloneqq \{g^k\ |\ |\textit{\textbf{x}}^k - \textit{\textbf{x}}_n(t)| \leq \delta\}$ as the target candidate grids to uphold the algorithms's scalability relative to the field size. Finally, $r_n$ selects $g^k$ with the largest utility as its target by:

\begin{subequations}
  \begin{equation}
    \begin{split}
      U_n^k(t) = \alpha_n^k(t)\frac{i_n^k(t) + \Delta _n^k(t)}{\Delta_n^k(t)}
    \end{split}
  \end{equation}      
  \begin{equation}
    \alpha_n^k(t) = 
      \begin{cases}
        \displaystyle(1 - \frac{\epsilon_n(t) - E}{\epsilon_{\mathrm{max}}})\ \ (\mathrm{if} \ \ \epsilon_n(t) \geq E,\ |\textit{\textbf{x}}^k| \geq |\textit{\textbf{x}}_n(t)|)
        \\
        \\
        \displaystyle(1 + \frac{\epsilon_n(t) - E}{\epsilon_{\mathrm{max}}})\ \ (\mathrm{elseif}\ \ \epsilon_n(t) \geq E,\ |\textit{\textbf{x}}^k| < |\textit{\textbf{x}}_n(t)|)
        \\
        \\
        \displaystyle(1 + \frac{E - \epsilon_n(t)}{\epsilon_{\mathrm{max}}})\ \ (\mathrm{elseif}\ \ \epsilon_n(t) < E,\ |\textit{\textbf{x}}^k| \geq |\textit{\textbf{x}}_n(t)|)
        \\
        \\
        \displaystyle(1 - \frac{E - \epsilon_n(t)}{\epsilon_{\mathrm{max}}})\ \ (\mathrm{elseif}\ \ \epsilon_n(t) < E,\ |\textit{\textbf{x}}^k| < |\textit{\textbf{x}}_n(t)|) 
    \end{cases}
    \label{eq2b}
  \end{equation}
\end{subequations}

\noindent where $\Delta_n^k(t)$ is the expected travel time for $r_n$ to $g^k$. $r_n$ amplifies or deminishes the utility of grids closer or farther to the base station by $\alpha_n(t)$, to modify the probability of moving towards the base station. \textit{E} is a constant parameter to switch these processes. To avoid target duplication with other robots, $r_n$ proactively updates the assumption for its target (line 11). This value will be shared with the surrounding robots by \textit{update\_assumption} at the next timestep, informing them of the reduced patrol need for this.

Function \textit{update\_epsilon} in Algorithm~\ref{alg2} updates $\epsilon_n(t)$ to convey the value to a robot that has been out of communication with the base station for a longer time. In the algorithm, $r_n$ compares $c_n^b(t)$ to $\forall r_m \in \mathcal{A}_n(t)$ and refresh $\epsilon_n(t)$. The process aims to offer robots equal communication opportunities with the base station.

\begin{algorithm}[tbp]
  \caption{Algorithm for updating $\epsilon_n(t)$ at time $t$}
  \label{alg2}
  \begin{algorithmic}[1]
    \Procedure{update\_epsilon}{$\epsilon_n(t), \forall c_m^b(t), \forall \epsilon_m(t) \mathrm{\ for\ } r_m \in \mathcal{A}_n(t)$}
      \If{$\mathcal{A}_n(t) \neq \emptyset$}
        \If{$\operatorname{min}(c_n^b(t),\ \forall c_m^b(t)) == c_n^b(t)$}
          \State $\text{\parbox[t]{\dimexpr\linewidth-\algorithmicindent}{$\displaystyle\epsilon_n(t) \leftarrow \operatorname{max}(\epsilon_n(t),\ \forall \epsilon_m(t)) + \eta_1 \sum_{m}{\epsilon_m(t)}\\ + \eta_2|\mathcal{A}_n(t)|$}}$
        \ElsIf{$\operatorname{max}(c_n^b(t),\ \forall c_m^b(t)) == c_n^b(t)$}
          \State $\epsilon_n(t) \leftarrow 0$
        \Else
          \State $\textit{is\_transferred} \leftarrow true$
          \ForAll{$r_m \in \mathcal{A}_n(t)$}
              \If{$c_n^b(t) < c_m^b(t)$ \textbf{\&\&} $\epsilon_m(t) > E_{\mathrm{th}}$}
                  \State $\epsilon_n(t) \leftarrow \operatorname{max}(\epsilon_n(t) + \eta_1 \epsilon_m(t),\ \forall\epsilon_m(t))$
                  \State $\textit{is\_transferred} \leftarrow false$
              \EndIf
          \EndFor
          \State $\epsilon_n(t) \leftarrow \epsilon_n(t) + \eta_2|\mathcal{A}_n(t)|$
          \If{$is\_transferred == true$}
              \State $\epsilon_n(t) \leftarrow 0$
          \EndIf
        \EndIf
      \EndIf
      \State send $\epsilon_n(t)$ and $c^b_n(t)$ to $\forall r_m \in \mathcal{A}_n(t)$
      \State \textbf{return} $\epsilon_n(t)$
    \EndProcedure
  \end{algorithmic}
\end{algorithm}

Per the LR algorithm, each robot independently selects its patrol target grid from its local area. Since such a reactive algorithm lacks a theoretical guarantee that all grids will eventually be patrolled, the comprehensiveness of patrolling is assessed through $I_W$, given a sufficiently prolonged mission duration.

\section{SIMULATIONS}\label{Sim}
\subsection{Compared Algorithm}
This study introduces three existing reactive patrol algorithms for comparison: CR~\cite{Machado2003}, ER~\cite{Yan2016}, DTAG and DTAP~\cite{Farinelli2017}. The study modified the algorithms for comparison by: (i)deploying a robot as a base station at the origin, (ii)limiting communication range as in LR, and (iii)enabling each robot to maintain $H_n(t)$ and share it with the base station. Given the absence of reactive algorithms that consider the base station, these algorithms are introduced into the study with a deployed base station serving as an SA center, while they do not have the intent to promote communication therewith. Parameters for each algorithm were adjusted through preliminary simulations.

\subsection{Configurations}
Table~\ref{tab1} presents simulation configurations. Each trial involves \textit{N} robots patrolling a $600 \times 600$ [m] field divided into $20 \times 20$ grids. At the beginning of each trial, robot $r_1$ is deployed as the base station at the origin, while others are deployed randomly. All idleness-related values are initialized to zero. During the mission, each robot selects its patrol target grid based on the current patrol algorithm. In the current setting, each robot takes $20-30$ seconds to travel from one grid to another. Accordingly, a robot spends $20-30$ seconds within a grid, assumedly completing its patrol for the grid in this period. Based on this assumption, the simulation abstracts the missions to a setting where the grid patrol completes when a robot comes within three meters of the center of the target grid. The study conducted 150 trials in total (10 trials, 5 algorithms, 3 different \textit{N}).

\begin{table}[tbp]
\caption{Parameters and configurations}
\label{tab1}
\begin{center}
  \begin{tabular}{cccl}
    \hline
    Symbol            & Value         & Unit    & Comment\\
    \hline
    $d_c$             & 180           & m       & Communication range\\
    $d_s$             & 90            & m       & Sensor range\\
    $\delta$          & 180           & m       & Consider range for next target\\
    $\epsilon_{\mathrm{max}}$  & 3000          & -       & -\\
    \textit{E}        & 500           & -       & -\\
    \textit{N}        & 4, 8, 12      & robots  & -\\
    \textit{K}        & $20 \times 20$& grids   & -\\
    -                 & $30 \times 30$& m       & grid size\\
    \textit{T}        & 40000         & sec     & duration of trial\\
    -                 & 10            & -       & \# of trials per \textit{N} per algorithm\\
    \hline
  \end{tabular}
\end{center}
\end{table}

\subsection{Results}
\begin{figure}[tpb]
  \centering
  \includegraphics[scale=0.31]{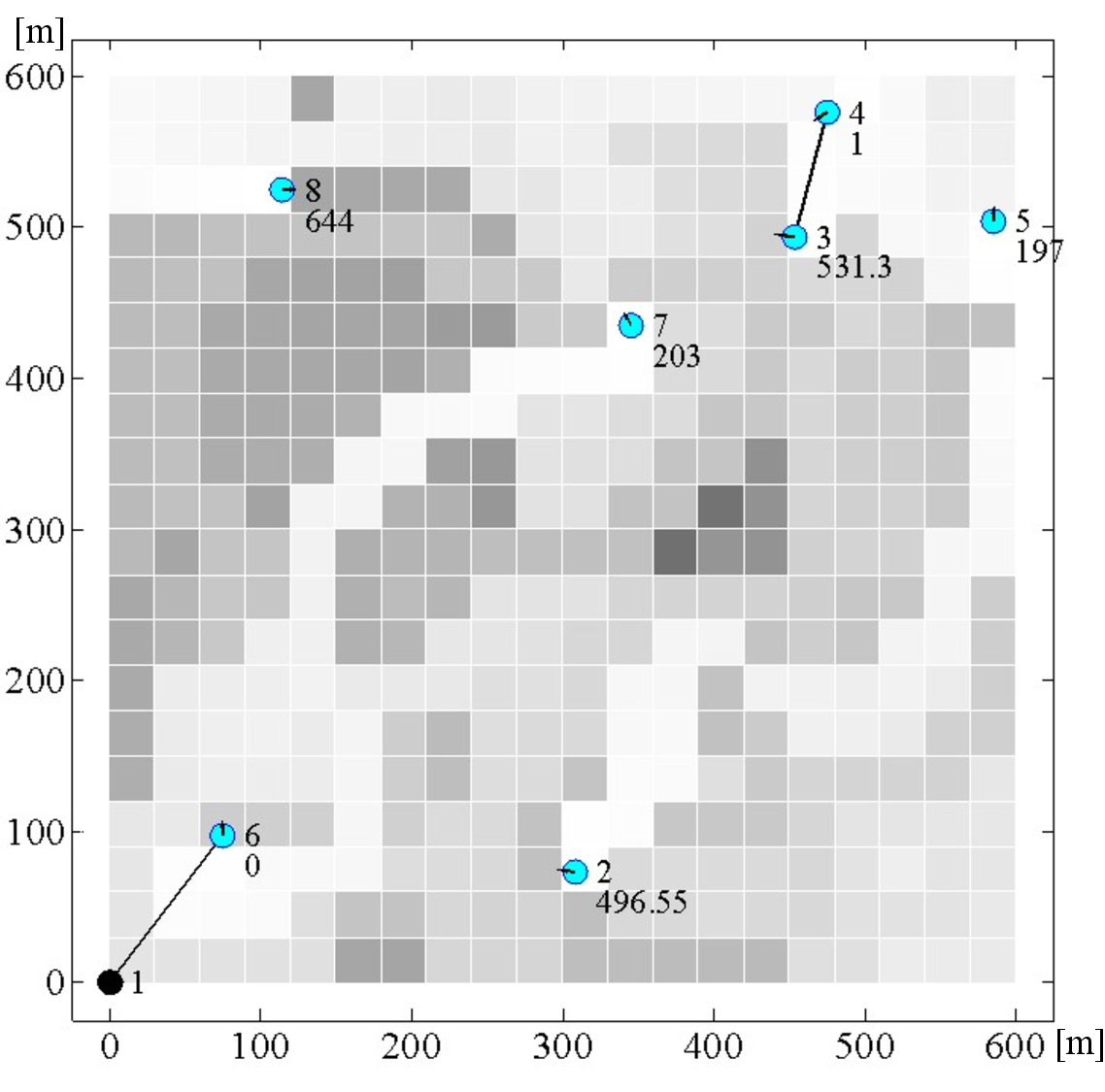}
  \caption{Screenshot of the simulated patrol mission ($N=8$). The black, cyan circles, and lines represent the base station, robots, and connections, respectively. Grid color indicates the ground truth of idleness: darker colors signify higher idleness.}
  \label{fig_capture}
\end{figure}

Figure~\ref{fig_capture} shows a simulation screenshot featuring eight robots. Grid color indicates idleness, with darker colors signifying higher idleness. Appended numbers to the robots represent ID (upper row) and $\epsilon_n(t)$ (lower row). The robots $r_3$ and $r_4$ are interconnected, thereby maintaining $\epsilon_4(t)$ at zero and aggregating to $\epsilon_3(t)$. 

\begin{figure}[tpb]
  \centering
  \includegraphics[scale=0.42]{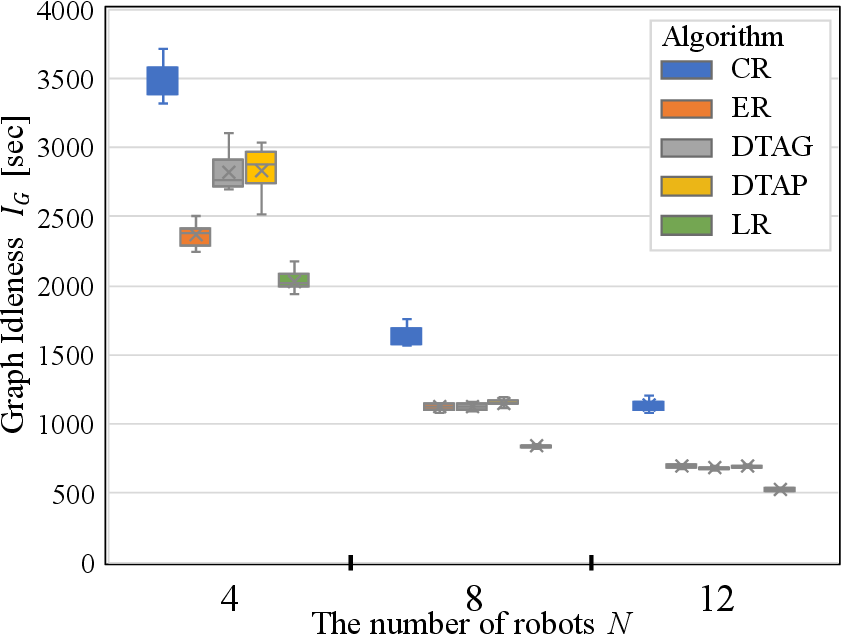}
  \caption{Box plot of Graph Idleness: $I_G$ by all ten trials}
  \label{fig_ig}
\end{figure}

\begin{figure}[tpb]
  \centering
  \includegraphics[scale=0.42]{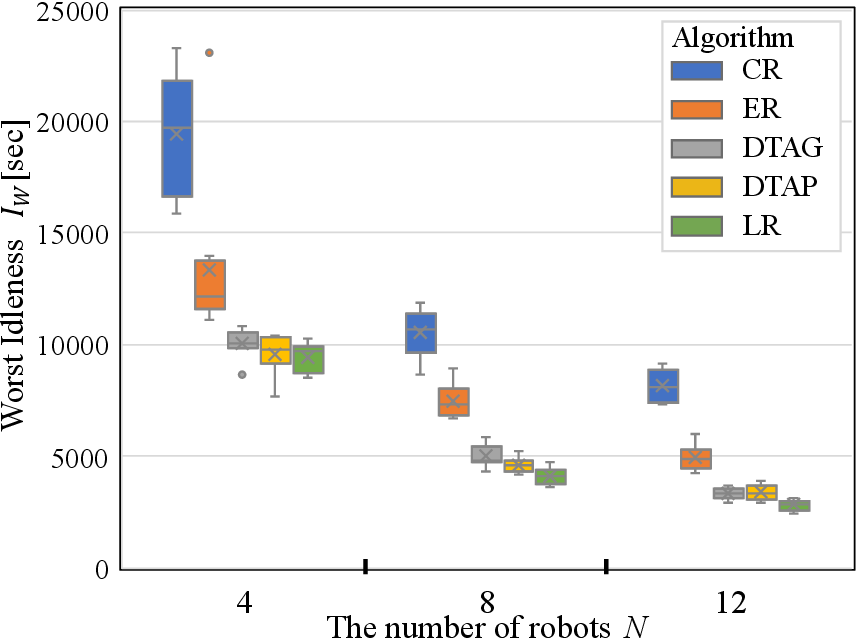}
  \caption{Box plot of Worst Idleness: $I_W$ by all ten trials}
  \label{fig_iw}
\end{figure}

\begin{figure}[tpb]
  \centering
  \includegraphics[scale=0.42]{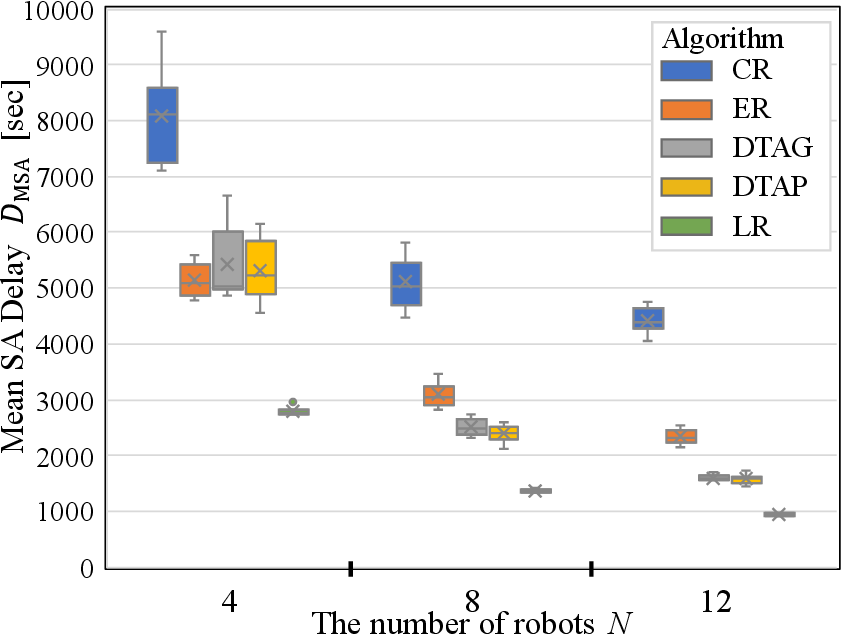}
  \caption{Box plot of Mean SA delay: $D_{\mathrm{MSA}}$ by all ten trials}
  \label{fig_dmsa}
\end{figure}

\begin{figure}[tpb]
  \centering
  \includegraphics[scale=0.42]{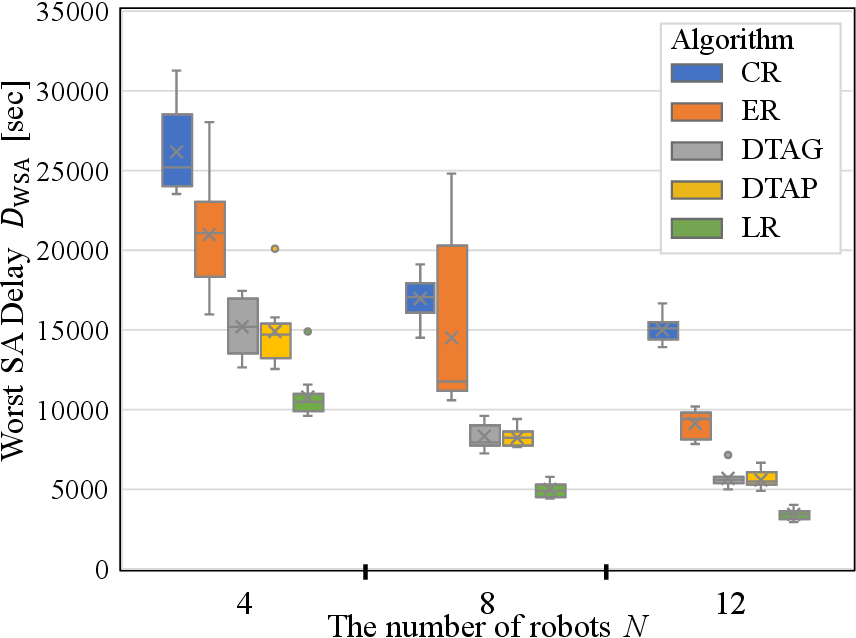}
  \caption{Box plot of Worst SA delay: $D_{\mathrm{WSA}}$ by all ten trials}
  \label{fig_dwsa}
\end{figure}

Figure~\ref{fig_ig} - \ref{fig_dwsa} show the performance by all the algorithms, with lower metrics indicating superior performance. An $I_W$ smaller than $T$ indicates that all the grids were patrolled at least once by the end of the missions. Figure~\ref{fig_ig} shows that LR generally excelled in terms of idleness. Though the robots under the compared algorithms operate independently of the base station, other characteristics, such as the design of information sharing, led to the inferior performance. On the other hand, Figure~\ref{fig_iw} shows that LR performed similarly to DTAG and DTAP in the comprehensiveness of patrolling. Since DTAG and DTAP account for all the patrol locations, the robots can prioritize grids with higher idleness. Regarding SA, Figure~\ref{fig_dmsa} and \ref{fig_dwsa} illustrate LR's superior performance over the other algorithms in providing timely information to the base station.

\section{CONCLUSION}\label{Conclude}
This study proposed a distributed multi-robot patrolling algorithm conscious of the base station’s situation awareness. The algorithm works according to individual robots' perceptions and communications with nearby robots. Though the proposed algorithm showed a similar performance to the existing algorithms in patrol comprehensiveness, it performed better in the base station's situation awareness evaluated by SA delay and the general patrol performance evaluated by the graph idleness.




\bibliographystyle{IEEEtran}
\bibliography{bib}

\begin{thebibliography}{10}
\providecommand{\url}[1]{#1}
\csname url@samestyle\endcsname
\providecommand{\newblock}{\relax}
\providecommand{\bibinfo}[2]{#2}
\providecommand{\BIBentrySTDinterwordspacing}{\spaceskip=0pt\relax}
\providecommand{\BIBentryALTinterwordstretchfactor}{4}
\providecommand{\BIBentryALTinterwordspacing}{\spaceskip=\fontdimen2\font plus
\BIBentryALTinterwordstretchfactor\fontdimen3\font minus
  \fontdimen4\font\relax}
\providecommand{\BIBforeignlanguage}[2]{{%
\expandafter\ifx\csname l@#1\endcsname\relax
\typeout{** WARNING: IEEEtran.bst: No hyphenation pattern has been}%
\typeout{** loaded for the language `#1'. Using the pattern for}%
\typeout{** the default language instead.}%
\else
\language=\csname l@#1\endcsname
\fi
#2}}
\providecommand{\BIBdecl}{\relax}
\BIBdecl

\bibitem{Schranz2020}
M.~Schranz, M.~Umlauft, M.~Sende, and W.~Elmenreich, ``{Swarm Robotic Behaviors
  and Current Applications},'' \emph{Frontiers in Robotics and AI}, vol.~7,
  no.~36, 2020.

\bibitem{Carrillo-Zapata2020}
D.~Carrillo-Zapata, E.~Milner, J.~Hird, G.~Tzoumas, P.~J. Vardanega,
  M.~Sooriyabandara, M.~Giuliani, A.~F. Winfield, and S.~Hauert, ``{Mutual
  Shaping in Swarm Robotics: User Studies in Fire and Rescue, Storage
  Organization, and Bridge Inspection},'' \emph{Frontiers in Robotics and AI},
  vol.~7, no.~53, apr 2020.

\bibitem{Gans2021}
N.~R. Gans and J.~G. Rogers, ``{Cooperative Multirobot Systems for Military
  Applications},'' \emph{Current Robotics Reports}, vol.~2, no.~1, pp.
  105--111, 2021.

\bibitem{Endsley1988}
M.~R. Endsley, ``{Design and Evaluation for Situation Awareness Enhancement},''
  \emph{Proceedings of the Human Factors Society Annual Meeting}, vol.~32,
  no.~2, pp. 97--101, oct 1988.

\bibitem{Banfi2015}
J.~Banfi, N.~Basilico, and F.~Amigoni, ``{Minimizing communication latency in
  multirobot situation-aware patrolling},'' in \emph{2015 IEEE/RSJ
  International Conference on Intelligent Robots and Systems (IROS)}, 2015, pp.
  616--622.

\bibitem{Scherer2020}
J.~Scherer and B.~Rinner, ``{Multi-Robot Patrolling with Sensing Idleness and
  Data Delay Objectives},'' \emph{Journal of Intelligent and Robotic Systems},
  vol.~99, no.~3, pp. 949--967, 2020.

\bibitem{Huang2019}
L.~Huang, M.~Zhou, K.~Hao, and E.~Hou, ``{A survey of multi-robot regular and
  adversarial patrolling},'' \emph{IEEE/CAA Journal of Automatica Sinica},
  vol.~6, no.~4, pp. 894--903, 2019.

\bibitem{Machado2003}
A.~Machado, G.~Ramalho, J.-D. Zucker, and A.~Drogoul, ``{Multi-agent
  Patrolling: An Empirical Analysis of Alternative Architectures},'' in
  \emph{Multi-Agent-Based Simulation II}, 2003, pp. 155--170.

\bibitem{Yan2016}
C.~Yan and T.~Zhang, ``{Multi-robot patrol: A distributed algorithm based on
  expected idleness},'' \emph{International Journal of Advanced Robotic
  Systems}, vol.~13, no.~6, pp. 1--12, nov 2016.

\bibitem{Farinelli2017}
A.~Farinelli, L.~Iocchi, and D.~Nardi, ``{Distributed on-line dynamic task
  assignment for multi-robot patrolling},'' \emph{Autonomous Robots}, vol.~41,
  no.~6, pp. 1321--1345, 2017.

\bibitem{Kobayashi2023a}
K.~Kobayashi, T.~Higuchi, and S.~Ueno, ``{Hierarchical and Distributed Patrol
  Strategy for Robotic Swarms with Continuous Connectivity},'' in
  \emph{Proceedings of the Joint Symposium of AROB-ISBC-SWARM2023}, 2023, pp.
  1491--1496.

\end{thebibliography}

\end{document}